\documentclass[conference]{IEEEtran}

\makeatletter
\def\ps@IEEEtitlepagestyle{%
  \def\@oddfoot{\mycopyrightnotice}%
  \def\@evenfoot{}%
}
\def\mycopyrightnotice{%
  {\footnotesize XXX-X-XXXX-XXXX-X/XX/\$XX.00~\copyright~20XX IEEE\hfill}
  \gdef\mycopyrightnotice{}
}
\usepackage{comment}

\usepackage{blindtext}
\usepackage{eso-pic}
\IEEEoverridecommandlockouts
\usepackage{cite}
\usepackage{amsmath,amssymb,amsfonts}
\usepackage{algorithmic}
\usepackage{graphicx}
\usepackage{caption}
\usepackage{subcaption}
\usepackage{textcomp}
\usepackage{xcolor}
\usepackage{multirow}
\def\BibTeX{{\rm B\kern-.05em{\sc i\kern-.025em b}\kern-.08em
    T\kern-.1667em\lower.7ex\hbox{E}\kern-.125emX}}
    
\usepackage{eso-pic}
\newcommand\AtPageUpperMyright[1]{\AtPageUpperLeft{%
 \put(\LenToUnit{0.17\paperwidth},\LenToUnit{-2cm}){%
     \parbox{0.9\textwidth}{\raggedleft\fontsize{8}{11}\selectfont #1}}%
 }}%
\newcommand{\conf}[1]{%
\AddToShipoutPictureBG*{%
\AtPageUpperMyright{#1}
}
}    
\usepackage{CJKutf8}
\usepackage[round]{natbib}
\begin{document}
\title{\vspace*{1cm} Knowledge-Prompted Estimator: A Novel Approach to Explainable Machine Translation Assessment}
\author{\IEEEauthorblockN{1\textsuperscript{st} Hao Yang}
\IEEEauthorblockA{\textit{2012 Labs} \\ Huawei Technologies CO., LTD \\
\textit{yanghao30@huawei.com}}
\and
\IEEEauthorblockN{2\textsuperscript{nd} Min Zhang }
\IEEEauthorblockA{\textit{2012 Labs}\\ Huawei Technologies CO., LTD \\
\textit{zhangmin186@huawei.com}}
\and
\IEEEauthorblockN{3\textsuperscript{rd} Shimin Tao}
\IEEEauthorblockA{\textit{2012 Labs}\\ Huawei Technologies CO., LTD \\
\textit{taoshimin@huawei.com}}
\and
\IEEEauthorblockN{4\textsuperscript{rd} Minghan Wang}
\IEEEauthorblockA{\textit{2012 Labs}\\ Huawei Technologies CO., LTD \\
\textit{wangminghan@huawei.com}}
\and
\IEEEauthorblockN{5\textsuperscript{rd} Daimeng Wei}
\IEEEauthorblockA{\textit{2012 Labs}\\ Huawei Technologies CO., LTD \\
\textit{weidaimeng@huawei.com}}
\and
\IEEEauthorblockN{6\textsuperscript{rd} Yanfei Jiang}
\IEEEauthorblockA{\textit{2012 Labs}\\ Huawei Technologies CO., LTD \\
\textit{jiangyanfei@huawei.com}}
}
\maketitle
\conf{\textit{  }}

\begin{abstract}
Cross-lingual Machine Translation (MT) quality estimation plays a crucial role in evaluating translation performance. GEMBA, the first MT quality assessment metric based on Large Language Models (LLMs), employs one-step prompting to achieve state-of-the-art (SOTA) in system-level MT quality estimation; however, it lacks segment-level analysis. In contrast, Chain-of-Thought (CoT) prompting outperforms one-step prompting by offering improved reasoning and explainability. In this paper, we introduce Knowledge-Prompted Estimator (KPE), a CoT prompting method that combines three one-step prompting techniques, including perplexity, token-level similarity, and sentence-level similarity. This method attains enhanced performance for segment-level estimation compared with previous deep learning models and one-step prompting approaches. Furthermore, supplementary experiments on word-level visualized alignment demonstrate that our KPE method significantly improves token alignment compared with earlier models and provides better interpretability for MT quality estimation. \footnote{Code will be released upon publication.}
\end{abstract}


\section{Introduction}
\label{sec:intro}


Large Language Models (LLMs), such as GPT-3 \citep{brown2020language}, ChatGPT \citep{kasirzadeh2023chatgpt}, GPT-4 \citep{openai2023gpt4}, and LLaMA \citep{touvron2023llama}, have been successfully validated in typical NLP scenarios, including question answering, search, summarization, and keyword extraction \citep{vilar2022prompting}. In the domain of multilingual NLP tasks, \citep{jiao2023chatgpt} \citep{hendy2023good} have demonstrated that utilizing a prompt-based approach, instead of fine-tuning models, can enable LLMs to perform machine translation tasks. This method has yielded impressive results for high-resource language pairs. However, for low-resource language pairs, the performance may be unsatisfactory due to insufficient training data or lower quality of available data.

\begin{figure}[t]
\centering
\resizebox{1\columnwidth}{!}{
    \includegraphics{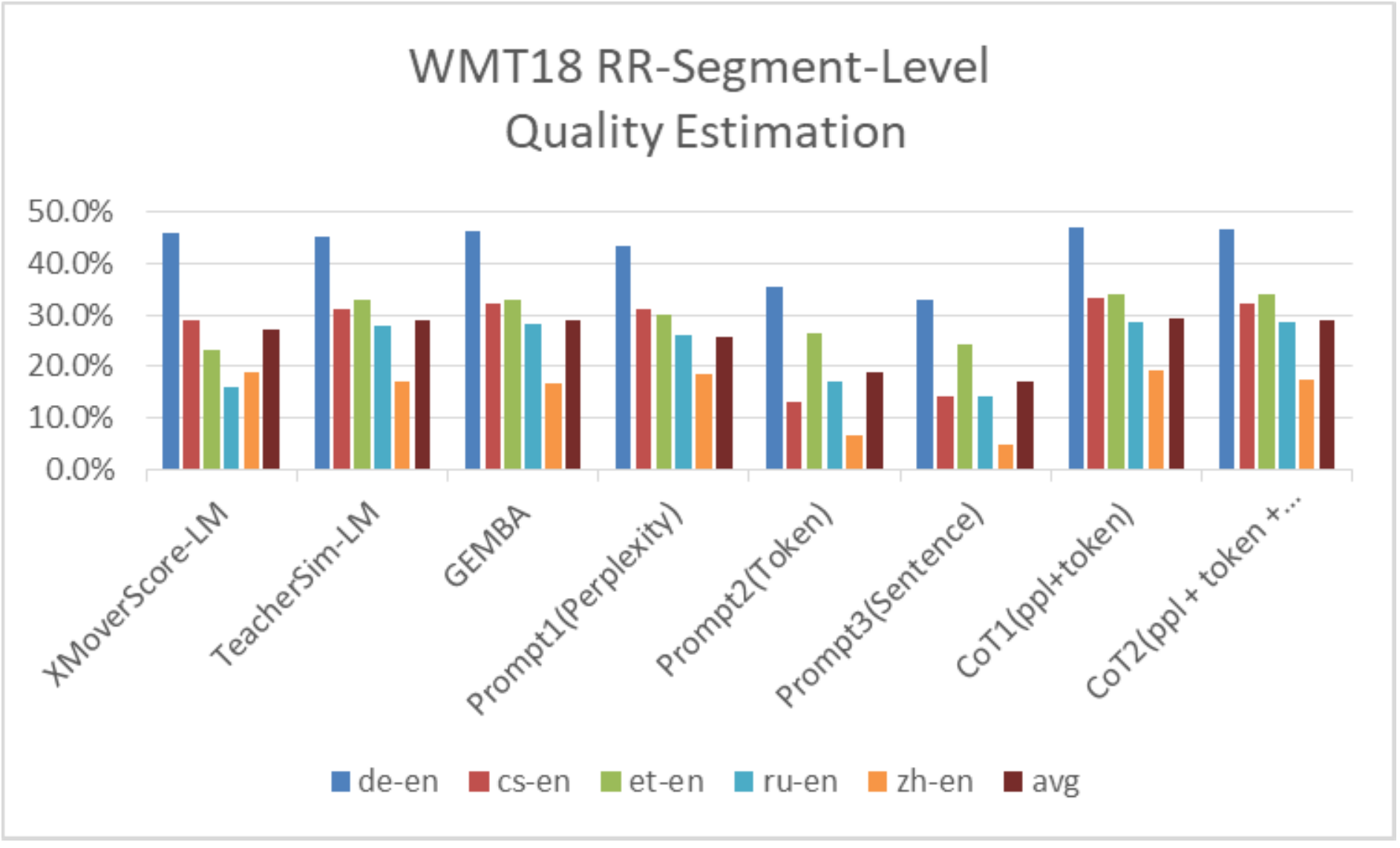}
}
\caption{The Kendall correlation of cross-lingual MT quality estimation for WMT18 segment-level metrics shows that the KPE CoT1 (CoT of perplexity and token) metric outperforms other metrics, including traditional deep learning metrics such as XMoverScore-LM and TeacherSim-LM, as well as LLM-based single-step metrics like GEMBA (perplexity, token, and sentence). Additionally, KPE CoT1 surpasses LLM-based CoT2, which is the CoT of perplexity, token, and sentence.}
\label{fig:self_constr_number}
\end{figure}

Going further, GEMBA \citep{kocmi2023large} explores the application of LLMs not only for translation tasks but also as translation quality estimators with one-step prompting. This approach targets two main scenarios: system-level quality estimation and segment-level quality estimation. By using one-step prompting, LLMs can assign scores to different systems or sentence pairs. Three scoring modes have been designed: scalar (0-100 points), 5-star (0-5 points), and 5-category (5-category classification). For evaluation, system-level assessment relies on accuracy as the metric, while segment-level assessment employs the Kendall ranking system as the evaluation metric. Results indicate that, under the prompting model and 5-category evaluation, state-of-the-art (SOTA) results are achieved at the system-level assessment. However, for segment-level quality assessment, LLMs still fall short compared with dedicated machine translation quality assessment models.

This study introduces two innovative considerations: (1) transitioning from single-dimensional evaluation to multi-dimensional evaluation. Machine translation Quality Estimation (QE) \citep{kocmi2021ship} now involves more than just assigning a single score; it adopts a multi-dimensional evaluation approach (MQM) \citep{rei2022comet}. This approach assesses fluency and accuracy separately. Fluency can be evaluated by having LLMs measure the perplexity of a sentence, while accuracy can be assessed by having LLMs evaluate sentence-level similarity \citep{yang2023teachersim} or word-level similarity \citep{zhang2023implicit}. (2) To implement multi-dimensional evaluation, the CoT prompting method for LLMs can be employed \citep{zhang2022automatic}\citep{fu2022complexity}. This approach guides LLMs to consider fluency first, followed by word-level and sentence-level accuracy, before finally combining 2-3 feature aspects to produce the best results. In the end, the study demonstrates that, in the WMT QE task, our KPE system achieves the best performance in 80\% of segment-level tasks. Additionally, in terms of interpretability \citep{tao2022crossqe}, token-level alignment exhibits better results.

In summary, this study can be characterized by the following key points:
\begin{itemize}
    \item Introducing Knowledge-Prompted Evaluator, including 3 one-step prompting evaluations for perplexity, token-level similarity, and sentence-level similarity; and 2 CoT prompting evaluations for perplexity-token prompting and perplexity-token-sent prompting.
    \item Experimental validation demonstrates that KPE achieves positive gains at each step of segment-level evaluation. Furthermore, one-step prompting is competitive and CoT1 prompting achieves SOTA segment-level evaluation performance, even better than CoT2 prompting.
    \item In terms of interpretability analysis, we compare the outcomes of BertScore, TeacherSim, and KPE. Our findings reveal that KPE-based token alignment exhibits substantially better accuracy and more discriminative power than previous results, thus offering enhanced interpretability capabilities.
\end{itemize}

\section{Related Work}
\label{sec:related}

\subsection{Machine Translation Quality Estimation}
Machine translation assessment can be classified into two categories based on the presence or absence of reference translations: quality evaluation as metrics, and Quality Estimation (QE) as metrics. Quality evaluation is a method that predicts the accuracy of machine translations based on a triplet of source text (src), machine translation output (mt), and reference translation (ref). In contrast, QE predicts the accuracy of machine translations solely based on the source text (src) and machine translation output (mt). Quality evaluation as metrics tend to have higher accuracy than QE. However, due to the need for human-provided reference translations for each source text, quality evaluation may be less efficient in practical applications. In contrast, QE as metrics, which does not require reference translations, has a wider range of applications despite its lower accuracy.

QE as metrics can be divided into two categories based on the evaluation methodology: system-level evaluation and segment-level evaluation.

System-level evaluation employs a calculation method similar to that of Learning-to-Rank (LTR), within two steps: For each system, it gets a system-level score with the source list (src list) and machine translation list (mt list). It then calculates the pairwise accuracy based on the machine-ranked and human labeled system scores. Segment-level evaluation, on the other hand, generates Relative Ranking (RR) segment-level data for all system pairs of source text (src) and machine translation (mt) using Direct Assessment (DA) or Multidimensional Quality Metrics (MQM). This data consists of triplets (src, mt1, mt2), indicating that the human evaluation score for mt1 is higher than that for mt2. Segment-level evaluation calculates correlations using the Kendall ranking system.
\begin{equation}
\begin{split}
QE = & fun(src, mt) \\
& - \text{segment-level $fun$, input (src, mt) return score} \\
   = & fun(src list, mt list) \\
& - \text{system-level $fun$, input (src list, mt list) return score}\\
\end{split}
\end{equation}

\subsection{Translation Quality Estimation Metrics}
Pairwise accuracy is the most commonly employed metric for system-level translation QE.
\begin{equation}
\begin{split}
Accuracy= &\frac{|sign(metric(\triangle) - sign(human(\triangle))|}{|all system pair|} \\
& - \text{$metric(\triangle)$  metric score pair difference} \\
& - \text{$human(\triangle)$ human labels  pair difference}\\
& - \text{$sign$  return 1 if first is better, else 0} \\
\end{split}
\end{equation}

where  Kendall's Tau \citep{callison2011findings}  is the most commonly employed metric for segment-level translation QE.

\begin{equation}
\begin{split}
\tau= &\frac{Concordant - Discordant}{Concordant+Discordant}  \\
& - \text{``Concordant'' - count of the set agrees with human labels} \\
& - \text{``Discordant'' - count of  the set disagrees with human labels}
\end{split}
\end{equation}

\begin{table}[!ht]
    \centering
    \begin{tabular}{|l|l|l|l|l|}
    \hline
        \multirow{5}{*}{Human} & & $s1 <s2$ & $s1=s2$ & $s1 < s2$ \\ \cline{2-5}
        & $s1 < s2$ & Concordant & Discordant & Discordant \\ \cline{2-5}
        & $s1=s2$ & $-$ & $-$ & $-$ \\ \cline{2-5}
        &  $s1 > s2$ & Discordant & Discordant & Concordant \\ \hline
    \end{tabular}
    \caption{Kendall metrics}
\end{table}

\subsection{LLM-based Quality Estimation Metrics}
Prompt engineering is the key component of success when it comes to using modern AI conversation tools and language models like ChatGPT and GPT-4. It is the art of crafting a statement or question that returns accurate and valuable results.

Different from traditional deep learning QE models, LLM QE or GEMBA is a prompt-tuning method based on LLMs. It consists of three steps: (1) finding an appropriate prompt template to complete the task, which includes a predefined response format; (2) filling the template with the source sentence, translation sentence, and other parameters; and (3) parsing results from the response.

The LLMs QE formula is defined as follows:
\begin{equation}
\begin{split}
    \text{LLMs QE} = &\text{fun(prompt, src list, mt list)} \\
    & - \text{$prompt$ , template string for quality estimation} \\
    & - \text{$src\ list$, source sentences} \\
    & - \text{$mt\ list$, translation sentences} \\
\end{split}
\end{equation}
The only difference between traditional QE and LLM QE lies in whether $prompts$ are used.

The generation of prompts can begin by asking counter-questions to LLMs. After obtaining multiple candidate templates, a suitable one can be manually selected and slightly modified. An example of a GEMBA prompt is as follows:

\emph{Classify the quality of machine translation into one of following classes: ``No meaning preserved'', ``Some meaning preserved, but not understandable'', ``Some meaning preserved and understandable'', ``Most meaning preserved, minor issues'', ``Perfect translation''.\\
source: ``{source\_seg}'' \\
machine translation: ``{target\_seg}'' \\
Class:}

\section{Proposed Approach}
\label{sec:method}
Prompt engineering can be categorized into two types: one-step prompting and Chain-of-Thought (CoT) prompting. One-step prompting is the most basic and straightforward prompt type, similar to zero-shot learning. Essentially, models like ChatGPT and GPT-3 complete the given prompt, which is the primary mechanism behind their functionality. One-step prompting works like a simple question-answer process, and even an initial statement can suffice, as the AI model will always attempt to complete it.

In contrast, CoT prompting not only provides AI with context for completion, but also offers a ``chain of thought'' process that demonstrates how the correct answer to a question should be reached. This type of prompting encourages reasoning and can even improve arithmetical results, which AI language models sometimes struggle with. Furthermore, due to its step-by-step reasoning approach, CoT prompting is much more explainable.

Our KPE tries to improve segment-level QE performance with CoT prompting. We first design three one-step prompting metrics, and then chain one-step prompting metrics as CoT metrics.

\subsection{KPE One-Step Prompting Metric}
Drawing inspiration from SOTA segment-level QE systems such as CometKiwi and TeacherSim, QE can be divided into three aspects, perplexity, token-level similarity and sentence-level similarity. We designed three one-step prompting metrics.

The one-step perplexity QE formula is defined as follows:

\begin{equation}
\begin{split}
    \text{Prompt1} = &\text{fun(perplexity\_prompt, mt)} \\
    \text{Prompt2} = &\text{fun(token\_sim\_prompt, src, mt)} \\
    \text{Prompt3} = &\text{fun(sent\_sim\_prompt, src, mt)} \\
\end{split}
\end{equation}
where $perplexity\_prompt$ is the prompt to estimate perplexity only based on mt, whereas $token\_sim\_prompt$ and $sent\_sim\_prompt$ are the prompts to estimate token level similarity and sentence level similarity.

\subsection{KPE CoT Prompting Metric}
KPE CoT prompting metrics are step-by-step prompting metrics based on three one-step prompting metrics.
\begin{equation}
\begin{split}
    \text{CoT1} = &\text{fun(comb\_prompt1, Prompt1, Prompt2)} \\
    \text{CoT2} = &\text{fun(comb\_prompt2, Prompt1, Prompt2, Prompt3)} \\
\end{split}
\end{equation}
where $comb\_prompt1$ is quality estimation based on perplexity and token-level similarity, whereas $comb\_prompt2$ is quality estimation plus sentence-level similarity.

\subsection{KPE Prompt Design}
\begin{figure*}[!h]
\centering
\includegraphics[width=0.95\linewidth]{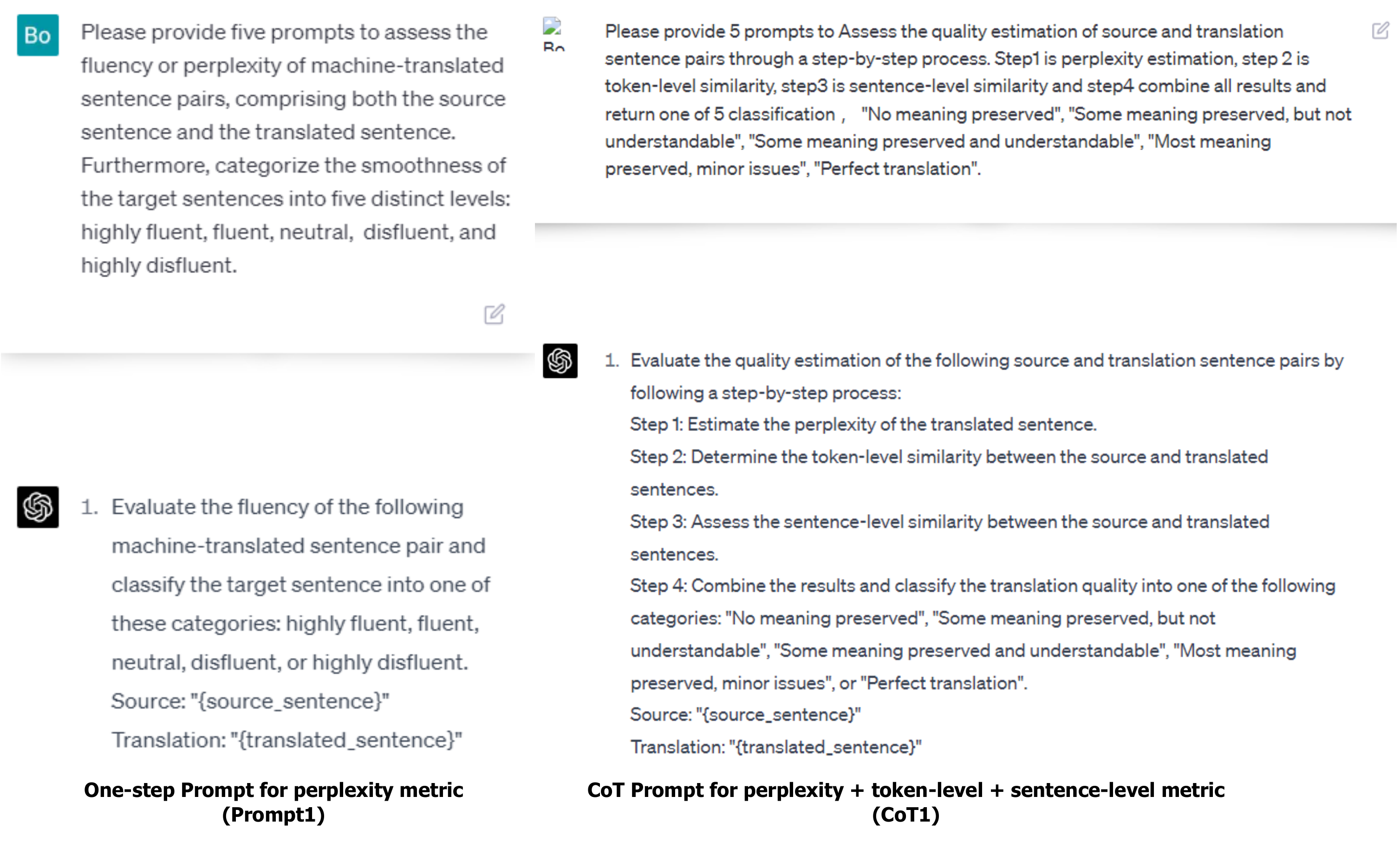}
\caption{One-step prompt for perplexity QE and CoT prompt for perplexity + token-level similarity and sentence-level similarity.}
\label{fig:prompt}
\end{figure*}

We follow the approach used in GEMBA for generating prompts. For the three one-step prompts and the two CoT prompts, we first have the system recommend five candidate prompt templates. We then manually select and edit these templates to create the optimal prompts. One-step prompts and CoT prompts are created like those in Figure \ref{fig:prompt}.

\section{Experiments}
\label{sec:experiment}

In this study, the proposed method is evaluated using the WMT18 metrics segment-level task data. For multilingual reference-free evaluation, the (source, candidate) pair is selected as a metric, similar to the QE approach.

\subsection{Datasets}
Our experiments employ the WMT18 news dataset as the evaluation dataset, containing 3,000 source sentences. The evaluation dataset also encompasses translated sentences from all teams that participated in WMT18. These target sentences are divided into 14 language pairs, such as DE-EN, FI-EN, and ZH-EN. Among them, DE-EN sentences are the largest in number (77,811), while CS-EN sentences are the smallest in number (5,110). Professional multilingual experts conduct pairwise comparisons on the target sentences before utilizing them as test data.
\begin{table}[h]
\caption{WMT18 segment-level to-En datasets.}
\centering\begin{tabular}{|l|c|c|c|c|c|c|c|}
\hline
Language Pair & cs-en & de-en & et-en & fi-en & ru-en & tr-en & zh-en \\
\hline
RR\_Systems            & 5     & 16    & 14    & 9     & 8     & 5     & 14    \\
\hline
Dev Datasize           & 5110  & 77811 & 56712 & 15648 & 10404 & 5525  & 33357 \\  \hline
\end{tabular}
    \label{tab:wmt18-rr-dataset}
\end{table}

\subsection{Models}
 The comparison systems comprise several strong baseline systems, including:
\begin{itemize}
    \item \textbf{M-BERT \citep{pires2019multilingual} and LASER \citep{artetxe2019massively} } as the multilingual pre-trained model. 
    \item \textbf{XMoverScore and XMoverScore+LM \citep{zhao2019moverscore} \citep{zhao-etal-2020-limitations}} based on token-level similarity.
    \item \textbf{TeacherSim and TeacherSim-LM \citep{yang2023teachersim} } as the strong baseline based on sentence-level similarity.
    \item \textbf{GEMBA \citep{kocmi2023large}} as the strong baseline based on one-step prompt and LLMs.
\end{itemize} 
In addition, we tested our method, which includes (1) three single-step approaches for perplexity, token similarity, and sentence similarity, and (2) two CoT methods: perplexity + token similarity; perplexity + token similarity + sentence similarity.

\subsection{Main Results}
Using multiple datasets such as the WMT18 to-En datasets, we compare the similarity of the source and translated sentences and estimate their quality by utilizing the Kendall rank correlation coefficient. The results reveal that: 
\begin{itemize}
    \item (1) One-step prompts achieve comparable performance, with Prompt1 (25.7\%), Prompt2 (18.8\%), and Prompt3 (17.1\%) performing better than traditional deep learning single models like M-BERT (11.0\%), LASER (15.0\%), XMoverScore (15.0\%), and TeacherSim (19.0\%); 
    \item (2) CoT1 (29.1\%) and CoT2 (28.9\%) outperform the combined method XMoverScore-LM (27.0\%) and LLM GEMBA (28.8\%);
    \item (3) CoT1 (29.1\%) achieves SOAT performance, surpassing CoT2 (28.9\%) and Teacher-LM (29.0\%), which indicates that increasing steps in prompting does not definitely improve the performance.
\end{itemize} 

\begin{table*}[h]
    \centering
    \caption{KPE result analysis for WMT18 segment-level to-En datasets.}
    \begin{tabular}{|l|l|l|l|l|l|l|l|l|l|}
    \hline
        \textbf{model} & \textbf{category} & \textbf{de-en} & \textbf{cs-en} & \textbf{et-en} & \textbf{fi-en} & \textbf{ru-en} & \textbf{zh-en} & \textbf{tr-en} & \textbf{avg} \\ \hline
        \textbf{M-BERT} & PLMs & 23.0\% & 1.0\% & 18.0\% & 12.0\% & 10.0\% & 8.0\% & 4.0\% & 11.0\% \\ \hline
        \textbf{LASER} & PLMs & 32.0\% & 7.0\% & 25.0\% & 16.0\% & 10.0\% & 6.0\% & 9.0\% & 15.0\% \\ \hline
        \textbf{XMoverScore} & Token & 28.0\% & 8.0\% & 21.0\% & 15.0\% & 15.0\% & 12.0\% & 9.0\% & 15.0\% \\ \hline
        \textbf{XMoverScore-LM} & Token + PLMs & 46.0\% & 29.0\% & 23.0\% & 32.0\% & 16.0\% & 19.0\% & 16.0\% & 27.0\% \\ \hline
        \textbf{TeacherSim} & Sentence & 17.0\% & 13.0\% & 17.0\% & 23.0\% & 26.0\% & 15.0\% & 21.0\% & 19.0\% \\ \hline
        \textbf{TeacherSim-LM} & Sentence+PLMs & 45.0\% & 31.0\% & 33.0\% & \textbf{24.0\%} & 28.0\% & 17.0\% & \textbf{22.0\%} & 29.0\% \\ \hline\hline
        \textbf{GEMBA} & One Step LLMs & 46.3\% & 32.1\% & 32.9\% & 23.9\% & 28.3\% & 16.7\% & 20.7\% & 28.8\% \\ \hline\hline
        \textbf{Prompt1(Perplexity)} & One Step LLMs & 43.2\% & 31.1\% & 29.9\% & 18.8\% & 25.9\% & 18.5\% & 12.3\% & 25.7\% \\ \hline
        \textbf{Prompt2(Token)} & One Step LLMs & 35.5\% & 13.0\% & 26.3\% & 16.4\% & 17.2\% & 6.7\% & 16.6\% & 18.8\% \\ \hline
        \textbf{Prompt3(Sentence)} & One Step LLMs & 32.9\% & 14.0\% & 24.3\% & 14.7\% & 14.2\% & 4.9\% & 14.7\% & 17.1\% \\ \hline
        \textbf{CoT1(ppl+token)} & CoT LLMs & \textbf{47.1\%} & \textbf{33.4\%} & \textbf{33.8\%} & 22.6\% & \textbf{28.7\%} & 
        \textbf{19.1\%} & 19.3\% & \textbf{29.1\%} \\ \hline
        \textbf{CoT2(ppl + token + sent)} & CoT LLMs & 46.7\% & 32.1\% & 33.8\% & 23.9\% & 28.4\% & 17.3\% & 20.4\% & 28.9\% \\ \hline
    \end{tabular}
\end{table*}
\begin{figure*}[h]
  \centering
  \begin{subfigure}[b]{0.3\textwidth}
    \includegraphics[width=\textwidth]{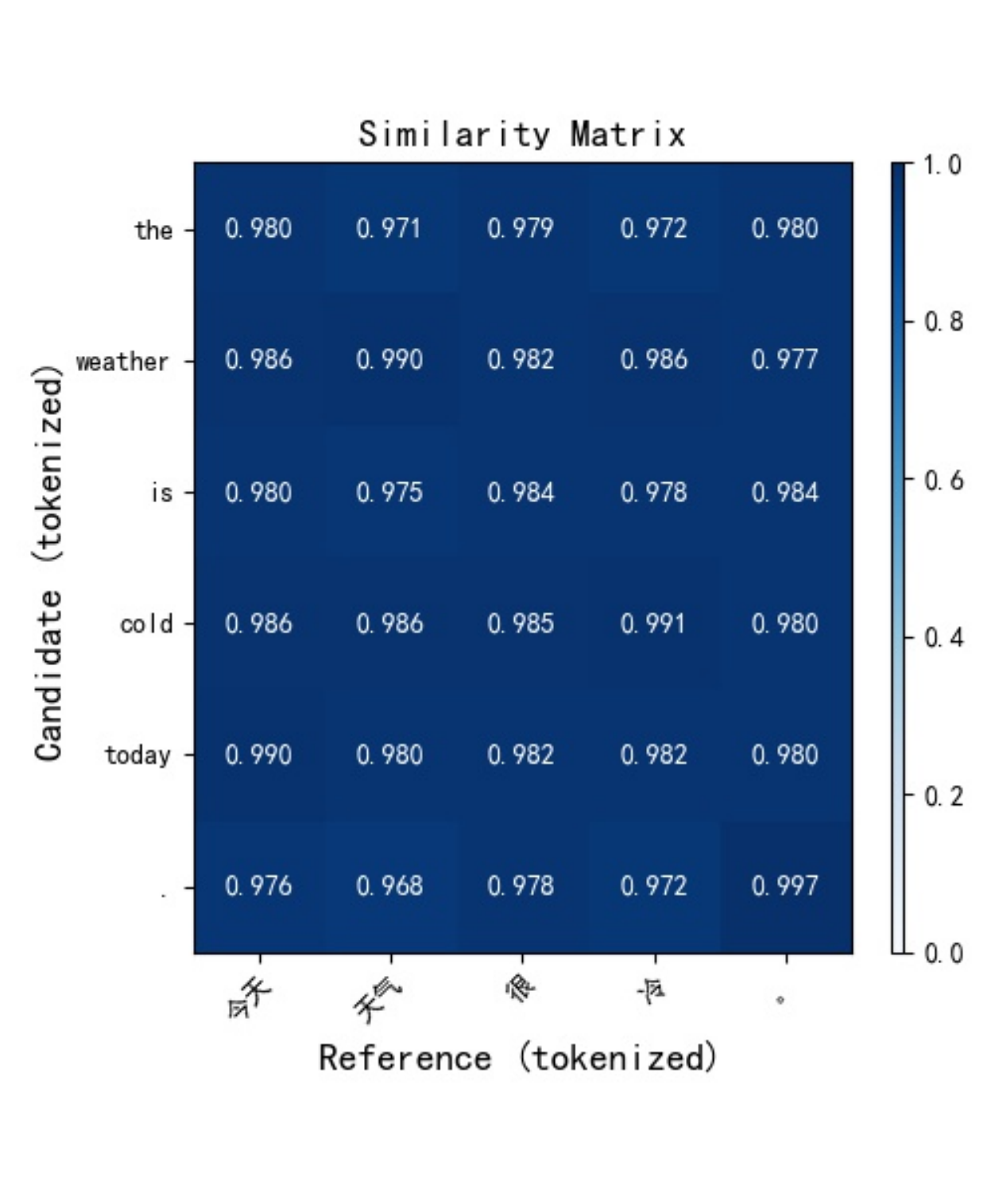}
    \caption{Token-level  alignment visualization based on BertScore over M-BERT}
    \label{fig:case_analysis1}
  \end{subfigure}
  \hfill
  \begin{subfigure}[b]{0.3\textwidth}
    \includegraphics[width=\textwidth]{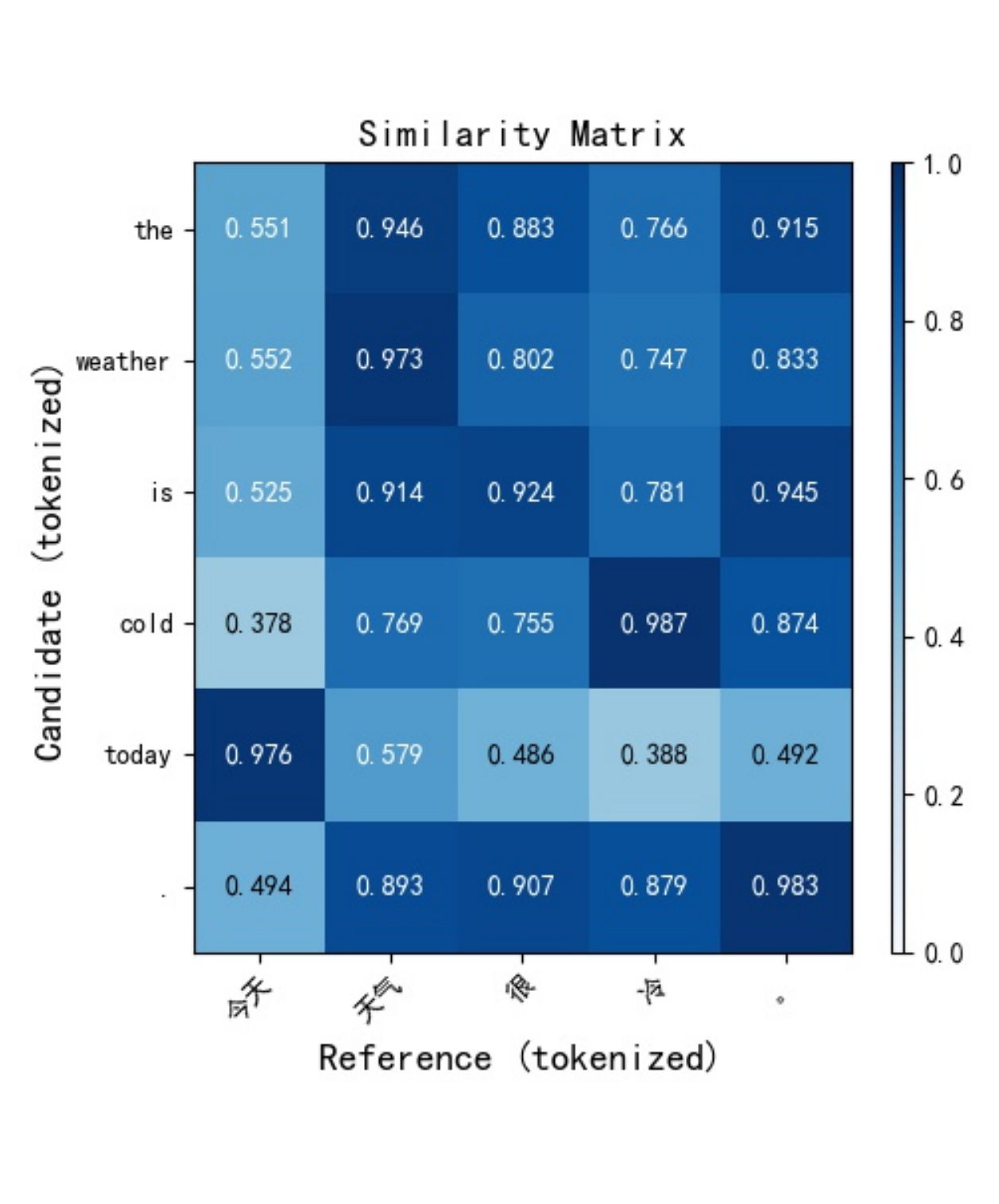}
    \caption{Token-level  alignment visualization based on BertScore over TeacherSim}
    \label{fig:case_analysis2}
  \end{subfigure}
  \hfill
  \begin{subfigure}[b]{0.3\textwidth}
    \includegraphics[width=\textwidth]{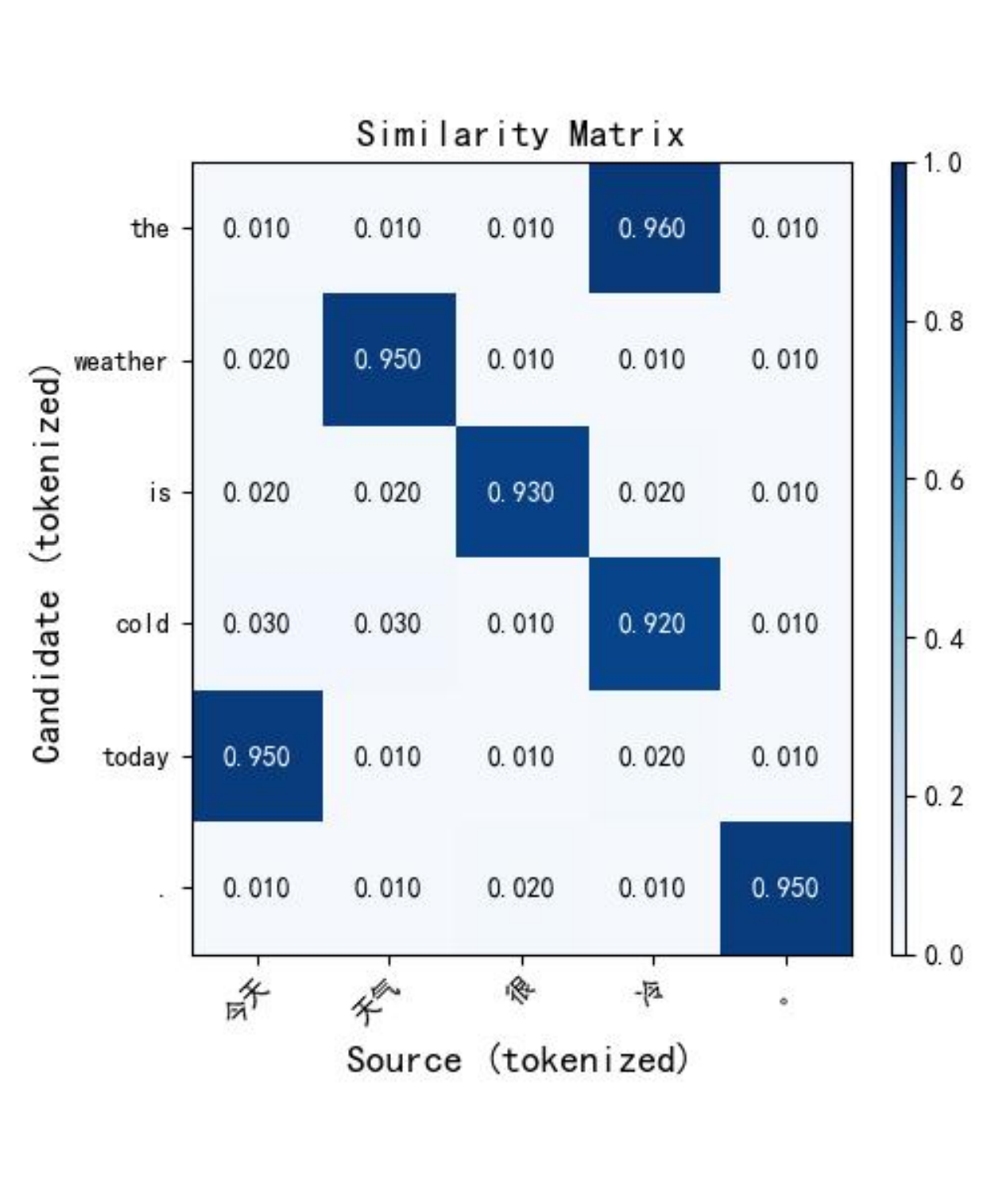}
    \caption{Token-level  alignment visualization based on KPE}
    \label{fig:llmqe}
  \end{subfigure}
  \caption{A comparison of explainable QE based on token-level alignment visualization for M-BERT, TeacherSim, and KPE demonstrates that KPE performs significantly better in terms of QE explanation.}
  \label{fig:images}
\end{figure*}

\subsection{Scorer Analysis}
Based on GEMBA's analysis, the scoring methods for QE can be divided into scalar scoring, 5-star scoring, and 5-category scoring. For segment-level evaluation, LLMs do not have high accuracy in scalar scoring, while 5-star and 5-category scoring methods perform better.

\begin{figure}[h]
\centering
\includegraphics[width=0.95\linewidth]{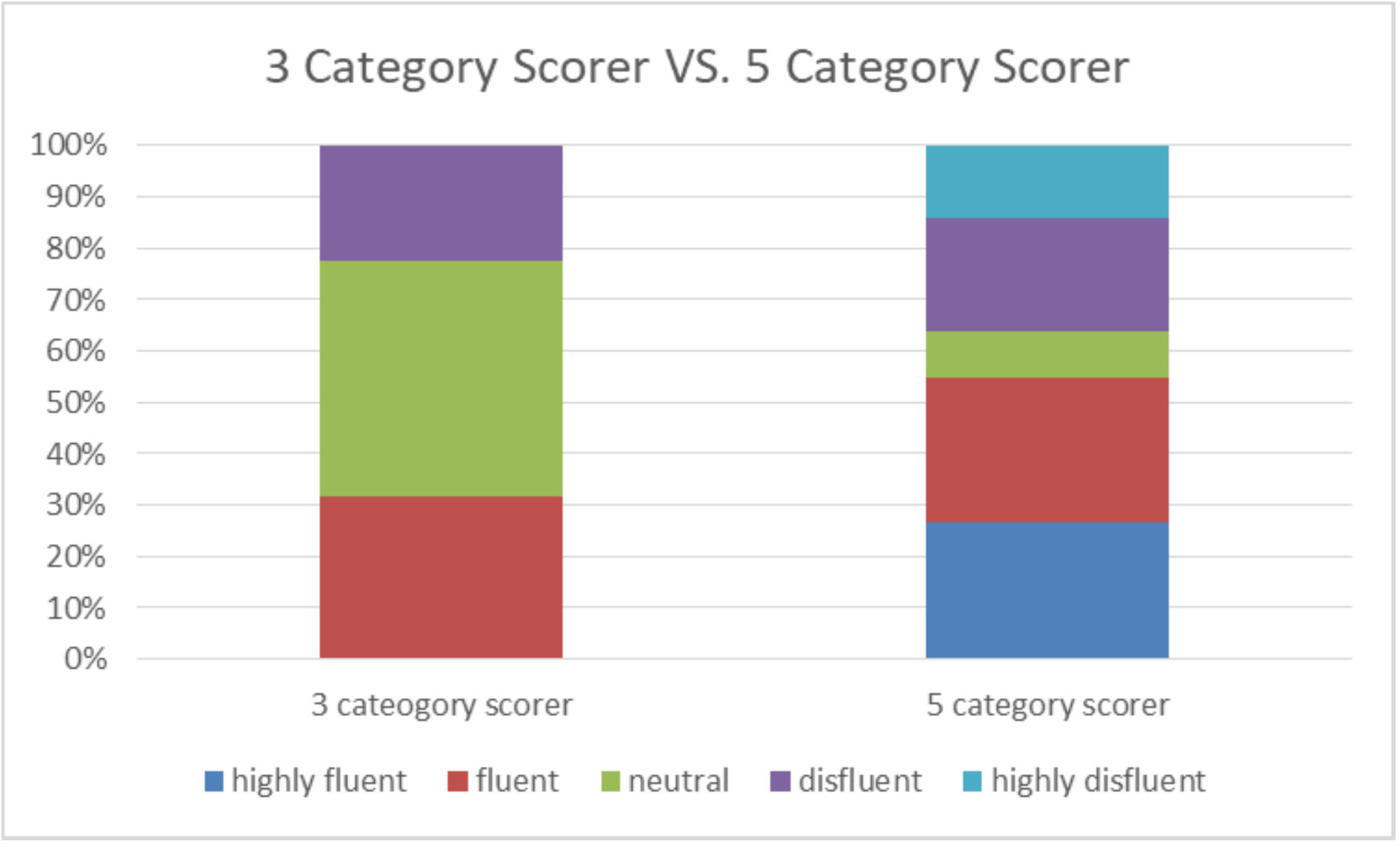}
  \caption{A comparison of the KPE 3-category scorer and the 5-class scorer shows that the 5-category scorer performs significantly better, primarily because fewer samples are labeled as neutral.}
  \label{fig:llm_scorers}
\end{figure}

In order to analyze the scoring capabilities of large models, we further examined the differences in scoring quality between 3-category and 5-category methods. We found that as machine translation quality improves, the 3-category model tends to gather more than 30\% in the neutral category, while the 5-category model has a much larger distinction, as shown in Figure \ref{fig:llm_scorers}.

\subsection{Explainable Analysis}
We also attempted to use large models for explainable analysis of QE in a similar manner to BERTScore or TeacherSim. We conducted visualization experiments for token-level alignment, calculating the similarity score for each token in the candidate sentence with each token in the reference sentence. This paper compares token-level alignment visualization for the multilingual pre-trained model XLM-R, the TeacherSim model, and KPE.

A typical case in Fig.\ref{fig:case_analysis1} shows that the similarity scores of all words in the multilingual pre-trained model M-BERT exceed 90\%, making it difficult to differentiate them. The token-level similarity distribution based on TeacherSim, as shown in Fig.\ref{fig:case_analysis2}, is considerably uneven. Although it is much more accurate than the multilingual pre-trained model, it still exhibits leaky probability and aligns to the last token, such as \textit{(``the'', ``.'')}, with an alignment probability of 91.5\%.

Compared with these models, KPE's token alignment avoids both the issue of alignment probability exceeding 90\% in XLM-R and the leaky probability phenomenon in TeacherSim. For instance, the alignment probability of \textit{("the", ".")} is 1\%, while that of \textit{(``.'', ``.'')} is 95\%, as shown in Fig\ref{fig:llmqe}.  The accuracy and explainability of KPE's token alignment are significantly better than those of previous models.


\section{Results}
\label{sec:results}


In this paper, we address the issue of low segment-level accuracy in quality estimation with single step prompts. We propose KPE, a QE method based on LLMs and CoT, which focuses on three core dimensions of quality estimation: perplexity, token similarity, and sentence similarity. Experimental results show that the CoT-based QE system outperforms both previous deep learning systems and single-step large model estimation methods in segment-level quality assessment. Moreover, KPE's token alignment visualization experiments demonstrate its clear superiority over multilingual pre-trained models and specialized sentence-level QE systems in terms of explainability.

As for future research directions, on the one hand, we will attempt to fine-tune LLMs of various sizes, such as 6B, 13B, or 65B LLaMA models, to explore the upper limit of large models in performing QE. On the other hand, for language pairs with lower Kendall coefficients, such as zh-en, we will try to incorporate knowledge from knowledge-graphs to improve the metrics for the relevant language pairs.

\bibliography{naacl22}
\bibstyle{naacl22}

\clearpage

\end{document}